\documentclass[conference,compsoc]{IEEEtran}
\usepackage{graphicx}
\usepackage{amsmath}
\usepackage{booktabs}
\usepackage{amsfonts}
\ifCLASSOPTIONcompsoc
  \usepackage[nocompress]{cite}
\else
  \usepackage{cite}
\fi
\ifCLASSINFOpdf
\else
\fi
\begin{document}
\title{SAM2-SGP: Enhancing SAM2 for Medical Image Segmentation via Support-Set Guided Prompting}

\author{\IEEEauthorblockN{Yang Xing}
\IEEEauthorblockA{J. Crayton Pruitt Family\\ Department of\\ Biomedical Engineering\\
University of Florida\\}
\and
\IEEEauthorblockN{Jiong Wu}
\IEEEauthorblockA{J. Crayton Pruitt Family\\ Department of\\ Biomedical Engineering\\
University of Florida\\}
\and
\IEEEauthorblockN{Yuheng Bu}
\IEEEauthorblockA{Department of Electrical \\ \& Computer Engineering\\
University of Florida\\}
\and
\IEEEauthorblockN{Kuang Gong}
\IEEEauthorblockA{J. Crayton Pruitt Family\\ Department of\\ Biomedical Engineering\\
University of Florida\\}}


% compsoc conferences compared to traditional conferences), use this
%Telephone: (800) 555--1212, Fax: (888) 555--1212}


\maketitle

\begin{abstract}
Although new vision foundation models such as Segment Anything Model 2 (SAM2) have significantly enhanced zero-shot image segmentation capabilities, reliance on human-provided prompts poses significant challenges in adapting SAM2 to medical image segmentation tasks. Moreover, SAM2's performance in medical image segmentation was limited by the domain shift issue, since it was originally trained on natural images and videos. To address these challenges, we proposed SAM2 with support-set guided prompting (SAM2-SGP), a framework that eliminated the need for manual prompts. The proposed model leveraged the memory mechanism of SAM2 to generate pseudo-masks using image–mask pairs from a support set via a Pseudo-mask Generation (PMG) module. We further introduced a novel Pseudo-mask Attention (PMA) module, which used these pseudo-masks to automatically generate bounding boxes and enhance localized feature extraction by guiding attention to relevant areas. Furthermore, a low-rank adaptation (LoRA) strategy was adopted to mitigate the domain shift issue. The proposed framework was evaluated on both 2D and 3D datasets across multiple medical imaging modalities, including fundus photography, X-ray, computed tomography (CT), magnetic resonance imaging (MRI), positron emission tomography (PET), and ultrasound. The results demonstrated a significant performance improvement over state-of-the-art models, such as nnUNet and SwinUNet, as well as foundation models, such as SAM2 and MedSAM2, underscoring the effectiveness of the proposed approach. Our code is publicly available at https://github.com/astlian9/SAM\_Support.
\end{abstract}


% \fi
\IEEEpeerreviewmaketitle

\begin{IEEEkeywords}
Auto-prompting, Fine-tuning, Foundation Model, Medical Image Segmentation, SAM2.
\end{IEEEkeywords}

\section{Introduction}
Vision foundation models have demonstrated strong zero-shot capabilities across various applications, including medical image segmentation~\cite{DBLP:journals/corr/abs-2108-07258}. Their impressive generalizability and few-shot learning capabilities make them attractive for adapting to downstream tasks, offering a more efficient alternative to training task-specific models from scratch. The segment anything model (SAM)~\cite{kirillov2023segment} is a recently developed visual foundation model designed for promptable image segmentation, pretrained on over 1 billion masks from 11 million natural images. Leveraging its large-scale training data and generalizable architecture, SAM exhibited strong zero-shot segmentation performance by using prompts as an extra input, such as a bounding box or positive and negative clicks, demonstrating exceptional generalization ability and establishing a new benchmark across various segmentation tasks~\cite{10386032, zhang2024surveysegmentmodelsam, shaharabany2023autosamadaptingsammedical}. Recent works also demonstrated the strong performance of the SAM model when applied to downstream medical image segmentation tasks ~\cite{ma2024segmentmedicalimagesvideos, wu2023medicalsamadapteradapting, yang2023sam3dsegment3dscenes, Gong_2024, CHEN2024103310, lei2024medlsamlocalizesegmentmodel}.

To extend these capabilities to more complex scenarios, the SAM2 model has been developed to expand the functionality of SAM to include video inputs~\cite{ravi2024sam2segmentimages}. This extension enabled SAM2 to process temporal sequences of images, making it suitable for tasks that required the understanding of spatial continuity over multiple frames. Fine-tuning it on specific tasks ~\cite{bai2024revsam2promptsam2medical, he2025fewshotadaptationtrainingfreefoundation, zhao2024retrievalaugmentedfewshotmedicalimage} and directly evaluating it on few-shot segmentation~\cite{zhu2024medicalsam2segment, shen2025interactive3dmedicalimage, yan2024biomedicalsam2segment} are two ongoing research topics of SAM2 in medical image segmentation~\cite{he2024shortreviewevaluationsam2s}. Although these SAM2-based methods required minimal or no training data, they still had notable limitations. Firstly, their performance remained highly dependent on user-provided high-quality instructions. Due to this limitation, recent work on the SAM2 model focused mainly on interactive medical image segmentation. Furthermore, it was trained on natural images and videos and could face domain shift issues when applied to medical image segmentation tasks.
\begin{figure}
  \centering
   \includegraphics[width=\linewidth]{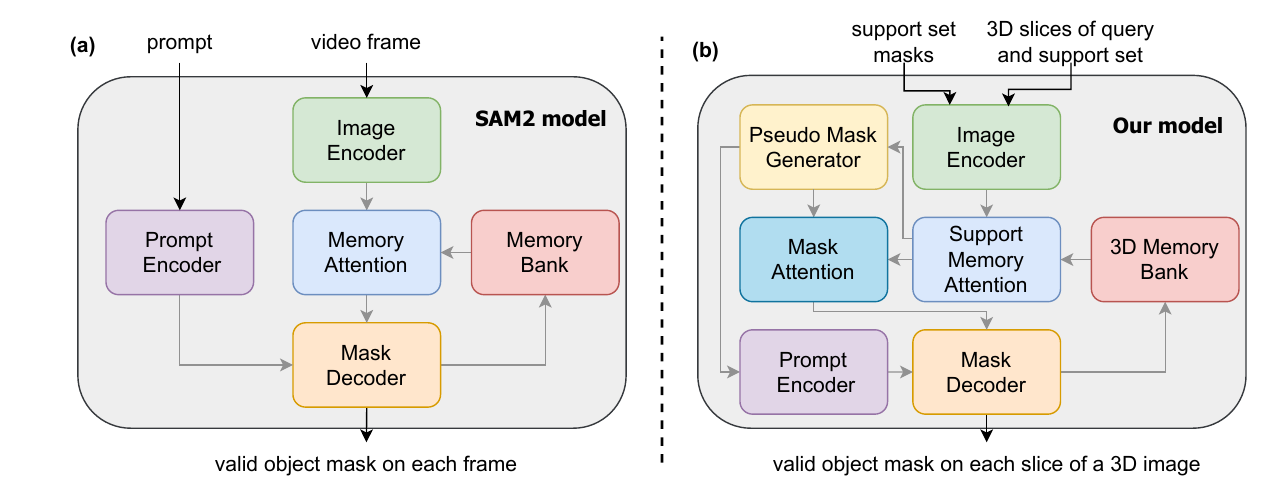}
   \caption{\small{A comparison between (a) SAM2 and (b) the proposed SAM2-SGP. More details of SAM2-SGP are shown in Fig. 2.
   }}
   \label{fig:comparison}
\end{figure}

To tackle the aforementioned limitations, we proposed a novel model, SAM2 with support-set guided prompting (SAM2-SGP), for medical image segmentation. By incorporating in-context learning with the memory mechanism of SAM2, SAM2-SGP could automatically generate high-quality prompts based on support sets. Specifically, the proposed SAM2-SGP model included a novel generator adapted from SAM2's memory mechanism to generate the pseudo-mask from image-mask pairs of the support set. These pseudo-masks were then used to compute bounding boxes, which could be used to generate prompt embeddings. Finally, image embeddings from the image encoder, pseudo-masks, and prompt embeddings from bounding boxes were subsequently processed by the pseudo-mask attention module for segmentation-map prediction.

Given the resemblance between video segmentation and 3D medical image segmentation, we also leveraged SAM2's memory bank and memory attention modules to enable 3D image segmentation by treating 3D images as a temporal sequence of 2D images.  A comparison between the proposed method and the original SAM2 model is shown in Fig.~\ref{fig:comparison}. Based on evaluations across multiple 2D and 3D medical imaging datasets, the proposed model consistently outperformed both fully supervised segmentation models and SAM2-based approaches. The major contributions of this work can be summarized as follows:
\begin{itemize}
\item We developed a prompt-free SAM2-based model for medical image segmentation. The proposed SAM2-SGP framework incorporated in-context learning with SAM2 and could achieve superior performance compared to other reference methods.
\item A pseudo-mask generation module adapted from SAM2's memory mechanism was introduced to generate pseudo-masks of query images based on the support set, enabling prompt generation without user interaction.
%\item We have developed a modified architecture of SAM2 so that it can be applied to medical tasks without the need for manual prompts. This refinement significantly boosts the segmentation performance of SAM 2. Remarkably, without any prompts, our automated segmentation consistently outperforms competitive SOTA foundation methods by a large margin.\\
\item A novel pseudo-mask attention module was introduced to generate the bounding-box prompts from pseudo-masks and improve localized feature extraction by using the pseudo-masks to guide attention to relevant areas.
% \item Low-rank adaptation (LoRA) was adopted to tackle the domain shift issue of SAM2 and enhance its feature extraction across different medical image modalities. 
\item The proposed model was evaluated on 2D and 3D datasets from different modalities, including optical, fundus photography, X-ray, ultrasound, CT, MRI and PET.
\end{itemize}

\section{Related works}
\label{sec:related_works}
\subsection{SAM2 model}
The SAM2 model inherited three core components from the original SAM architecture: an image encoder, a prompt encoder, and a mask decoder. The image encoder utilized a hierarchical Vision Transformer (ViT) backbone, Hiera~\cite{ryali2023hierahierarchicalvisiontransformer}, to extract multiscale feature embeddings from the input image. The prompt encoder processed various forms of user input, such as positive and negative clicks, bounding boxes, or dense masks, into a format suitable for segmentation guidance. The mask decoder refined segmentation predictions by integrating image and prompt features through bidirectional Transformer blocks. Building on this foundation, SAM2 extended the original SAM model to video segmentation by introducing a memory mechanism that enhanced temporal consistency across sequential video frames. This memory mechanism consisted of three key components: a memory encoder, a memory bank, and a memory-attention module. Unlike SAM, which processed each image independently, SAM2 encoded features and predicted masks from previous frames using the memory encoder and stored them in the memory bank. During inference, the memory-attention module retrieved relevant embeddings from the memory bank and integrated them with the current frame's features, refining segmentation predictions with temporal context. To support efficient memory usage, the memory encoder also downsampled predicted masks before storage. This structured memory-attention design enabled SAM2 to scale effectively while improving segmentation accuracy and robustness, particularly in challenging scenarios involving occlusion and object motion.
 
 Recent studies have increasingly explored the application of SAM2 to medical image segmentation. FS-MedSAM2~\cite{bai2024revsam2promptsam2medical} applied few-shot learning to evaluate SAM2 on the Synapse CT dataset~\cite{synapse2015}. FATE\_SAM~\cite{he2025fewshotadaptationtrainingfreefoundation} leveraged Dinov2~\cite{oquab2024dinov2learningrobustvisual} to select the most similar samples for few-shot learning-based image segmentation. Similarly, Zhao et al.~\cite{zhao2024retrievalaugmentedfewshotmedicalimage} adopted DINOv2 to construct support sets and evaluated SAM2 on the STARE dataset. In terms of fine-tuning, current efforts remained focused on prompt-based segmentation, which typically relied on high-quality human-provided prompts. Notable examples include MedSAM2~\cite{zhu2024medicalsam2segment}, SAM2\_med\_3D~\cite{shen2025interactive3dmedicalimage}, RevSAM2~\cite{sun2024vrpsamsamvisualreference}, and BioSAM2~\cite{yan2024biomedicalsam2segment}. In contrast to these approaches, the proposed SAM2-SGP model leveraged in-context learning in conjunction with SAM2’s memory mechanism to enable automatic prompting for medical image segmentation, reducing the reliance on manual input.

 \begin{figure*}[t]
  \centering
   \includegraphics[width=0.9\linewidth]{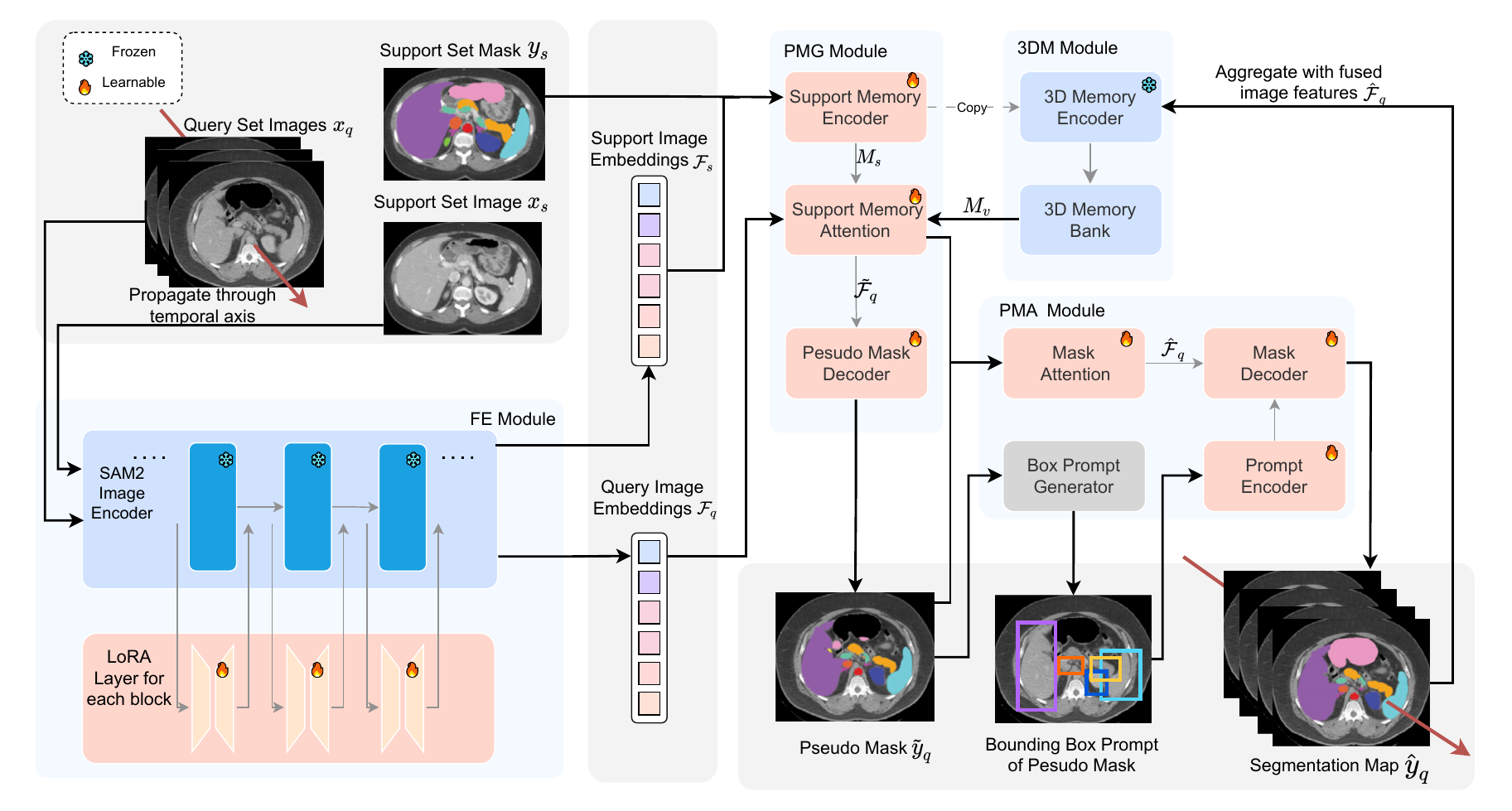}
   \caption{\small{Diagram of the proposed framework. It contained four modules: (a) feature extraction (FE) module, (b) pseudo-mask generation (PMG) module, (c) pseudo-mask attention (PMA) module, and (d) 3D memory encoding (3DM) module. The 3D memory encoding module was only enabled when processing 3D datasets. }}
   \label{fig:SAM2_support}
\end{figure*}

\subsection{In-context learning}
In-context learning (ICL) is a paradigm in which a model leverages a small set of examples (support set) provided during inference to make predictions on new inputs (query set). Unlike traditional fine-tuning, where models undergo explicit weight updates, ICL allows models to dynamically adapt by conditioning on context examples, making it particularly useful for few-shot learning scenarios. In medical image segmentation, ICL has been explored in Transformer-based architectures~\cite{shi2022densecrossqueryandsupportattentionweighted}, where models were adapted to new tasks by conditioning on input queries and task-specific demonstrations. Universeg~\cite{butoi2023universeguniversalmedicalimage} trained a CNN with inputs from both the query set and the support set so that the support set could be seen as task-specific demonstrations for predicting the query set. ICL-SAM~\cite{pmlr-v250-hu24a} followed Universeg’s design and combined the ICL model and the SAM model for segmentation tasks.  We observed that ICL conditioned query features on support set features, similar to how SAM2 conditioned current frame features on features from previous frames. Inspired by this connection, we proposed SAM2-SGP, which incorporated ICL into the SAM2 model. In SAM2-SGP, a subset of the training data was designated as the support set, which guided the generation of pseudo-masks used as the prompt.

\section{Methodology}
\label{sec:methodology}
\subsection{Problem definition}
%Our goal is to design one single model adapted to both 2D and 3D medical image segmentation. 
Consider a segmentation task with a training set of $N$ image-label pairs, $\boldsymbol{S}_{train}=\{(x^{train}_i,y^{train}_i)\}^N_{i=1}$, and a testing set of $M$ image-label pairs, $\boldsymbol{S}_{test}=\{(x^{test}_j,y^{test}_j)\}^M_{j=1}$. A standard strategy in deep learning-based image segmentation is to learn a parametric function $\hat{y}=f_{\boldsymbol{\theta}}(x)$ to directly predict the segmentation map based on the input image $x$, where $f_{\boldsymbol{\theta}}(\cdot)$ indicates the network, and $\boldsymbol{\theta}$ denotes the network parameters. In our proposed framework, we randomly divided the training set into a support set $\boldsymbol{S}_{sup} = \{(x^{train}_k,y^{train}_k)\}^{\mathcal{D}}_{k=1}$ and a query set $\boldsymbol{S}_{qry} = \boldsymbol{S}_{train}\setminus{\boldsymbol{S}_{sup}}$, where $\mathcal{D}=|\boldsymbol{S}_{sup}|$ and $|\boldsymbol{S}_{qry}|\gg\mathcal{D}$. 
 For each query image $x_q \in \boldsymbol{S}_{qry}$,we selected a subset $\tilde{\boldsymbol{S}}_{{sup}} \subset \boldsymbol{S}_{{sup}}$ consisting of the $K$ most similar samples. Our goal was to train a function $\hat{y}_q = f_{\boldsymbol{\theta}}(x_q, \tilde{\boldsymbol{S}}_{{sup}})$ that could accurately predict the segmentation map $\hat{y}_q$ based on the input image $x_q$ and its corresponding support subset $\tilde{\boldsymbol{S}}_{{sup}}$.
 
For 3D image segmentation, we aimed to fully exploit spatial context by leveraging the memory mechanism of SAM2 to improve prediction quality and ensure inter-slice coherence. Given a 3D image $x_q$ from $\boldsymbol{S}_{qry}$ with dimensions $D\times{H}\times{W}$, where $D$ was the number of 2D slices and $H\times{W}$ was the spatial size of each slice, the segmentation task involved training a parametric function $\hat{y}_q^i = f_{\boldsymbol{\theta}}(x_q^i, \hat{\boldsymbol{S}}_{{sup}})$ to predict the segmentation map for slice $i$, where $i \in {1, 2, \dots, D}$. 
 Here, the dynamic support set $\hat{\boldsymbol{S}}_{{sup}}$ was defined as $\hat{\boldsymbol{S}}_{sup}=\tilde{\boldsymbol{S}}_{sup}\cup\{(x_q^j,\hat{y}_q^j)\}_{j=1}^{i-1}$, which combined: (1) the $K$ most similar 2D slices from the static support subset $\tilde{\boldsymbol{S}}_{{sup}}$, and (2) the previously seen slices of $x_q$ along with their predicted segmentation maps. This formulation enabled the model to condition each slice’s prediction not only on external support examples but also on the evolving context of the preceding slices within the same volume.
 \subsection{Overall architecture}
As shown in Fig.~\ref{fig:SAM2_support}, the proposed model comprised four key components: the feature extraction (FE) module, the pseudo-mask generation (PMG) module, the 3D memory encoding (3DM) module, and the pseudo-mask attention (PMA) module. The feature extraction module processed images from both the query and support sets, extracting multiscale features. The pseudo-mask generation module encoded the extracted image embeddings of the support set with the corresponding masks to generate support memories. These support memories were then fused with the query image embeddings and passed through a decoder to produce a pseudo-mask for the query image. The pseudo-mask attention module leveraged this pseudo-mask to generate a bounding box prompt, applied mask attention between the pseudo-mask and query embeddings, and outputed a segmentation map based on the box prompt, pseudo-mask and query embeddings. For the 3D memory encoding module, the prediction of the current slice was encoded to obtain the image embeddings and appended to the support memory to predict the pseudo-mask of the next slice. The details of each module are explained below.
\subsection{Feature extracting module}
The feature extraction module was supplied with images from both the query and support sets to extract multiscale features. The architecture was based on the SAM2 image encoder, which utilized Hiera~\cite{ryali2023hierahierarchicalvisiontransformer} as the backbone. Given that the image encoder constituted approximately 70\% of SAM2’s total parameters, we integrated Low-Rank Adaptation (LoRA)~\cite{hu2021loralowrankadaptationlarge} layers to enable efficient fine-tuning. LoRA achieved this by decomposing weight updates into low-rank matrices, allowing adaptation with a reduced number of trainable parameters while preserving the pre-trained model’s knowledge. During the training process, the weight of the SAM2 image encoder remained frozen and only the LoRA layers were fine-tuned. The image embeddings of the query set $\mathcal{F}_{q}$ and the support set $\mathcal{F}_{s}$ were denoted as
\begin{equation}
\begin{aligned}
\mathcal{F}_{q} = \mathcal{E}(x_q),\\
\mathcal{F}_{s} = \mathcal{E}(x_s),\\
\end{aligned}
    \label{eq:image_encoder_support}
\end{equation}
where $\mathcal{E}(\cdot)$ was the SAM2 image encoder with LoRA layers.

\begin{figure*}
  \centering
  \includegraphics[width=0.95\linewidth]{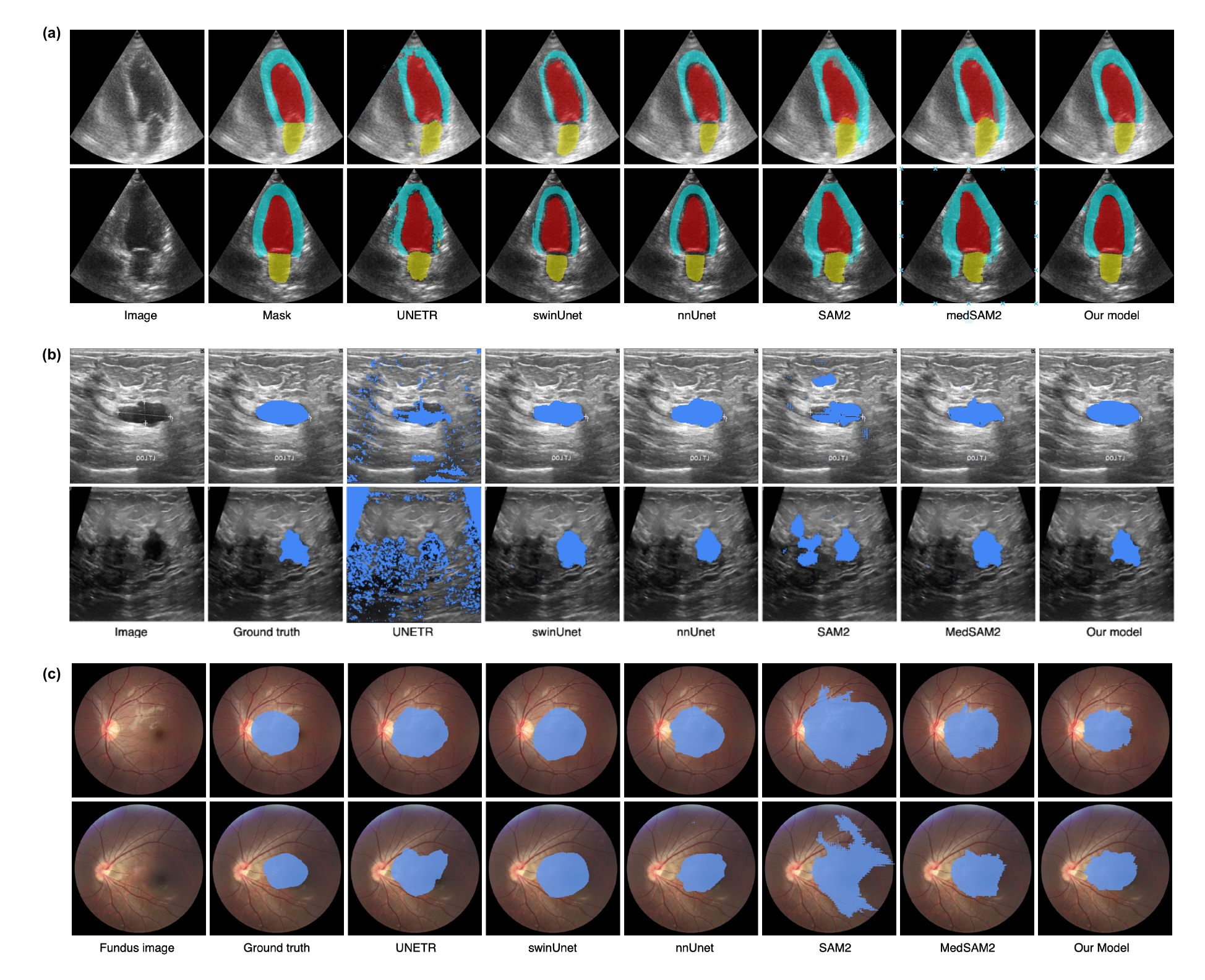}
  \caption{\small{Examples of segmentation results from the CAMUS, BUSI, and REFUGE datasets. For each sub-figure, the first column shows the image, and subsequent columns present results from ground truth and 6 comparison methods. The two rows represent two different cases. (a) Segmentation results of the CAMUS dataset. The left ventricle, the atrium, and the myocardium were labelled in red, yellow, and cyan, respectively. (b) Segmentation results of the BUSI dataset. Tumors were segmented out in blue. (c) Segmentation results of the REFUGE dataset. Optic disc was labeled in blue.}}
  \label{fig:2Dvisualization}
\end{figure*}

\subsection{Pseudo-mask generation module}
SAM2's memory components were originally developed to capture relationships between visual entities across images (i.e., tracking moving muscles in a ultrasound video), providing a robust mechanism for cross-image feature correlation. Inspired by in-context learning, this architectural capability was integrated into our proposed model to implement cross-image attention that enabled extraction of relevant query-image information conditioned on the corresponding features in the support set. To leverage this capability, the PMG module adapted SAM2 model's memory encoder and memory attention blocks to generate initial pseudo-masks that guided subsequent processing stages. This module had three components: the support-memory encoder, the support-memory attention block, and the pseudo-mask decoder. The support-memory encoder, adapted from the SAM2 memory encoder, extracted memory features from the support set. It combined multiscale features of the support set, $\mathcal{F}_{s}$, with segmentation masks from the support set, $y_s$, through
\begin{equation}
    \mathcal{M}_s = \phi(y_s)+\mathcal{F}_s,
    \label{eq:support_memory_encoder}
\end{equation}
where $\phi(\cdot)$ was the convolutional module downsampling $y_s$ to the same dimension as $\mathcal{F}_s$. The support-memory attention block aligned the encoded support-set features, $\mathcal{M}_s$, with query-set features, $\mathcal{F}_q$, to produce query-specific feature representation $\tilde{\mathcal{F}_q}$, which could be denoted as
\begin{equation}
\begin{aligned}
    &\tilde{\mathcal{F}_q} = MA(\mathcal{M}_s, \mathcal{F}_q)=CA(\mathcal{M}_s, SA(\mathcal{F}_q)),\\
    &CA(x_1,x_2)=softmax(\frac{Q_1K_2^T}{\sqrt{d}}V_2+V_2),\\
    &Q_1 = W^Q x_1,\quad K_2 = W^K x_2,\quad V_2 = W^V x_2,\\
    &SA(x)=softmax(\frac{QK^T}{\sqrt{d}}V+V),\\
    &Q = W^Q x,\quad K = W^K x,\quad V = W^V x,
\end{aligned}
    \label{eq:memory_attention}
\end{equation}
where $MA(\cdot)$ was the memory attention module, which first applied a self-attention block with residual paths, $SA(\cdot)$, and then further went through a cross-attention block with residual paths, $CA(\cdot)$. $W^Q, W^K, W^V$ denoted learnable projection matrices in the corresponding attention blocks that transformed input features into $Q$ (query), $K$ (key), $V$ (value), respectively. The pseudo-mask decoder processed the encoded query set features $\tilde{\mathcal{F}_q}$ to generate a pseudo-mask $\tilde{y}_q$ for the query image using an empty prompt input, which could be written as
\begin{equation}
    \tilde{y}_q = \mathcal{D}(\tilde{\mathcal{F}_q}, \mathcal{PE}(None)),
    \label{eq:pseudo_mask_decoder}
\end{equation}
where $\mathcal{D}(\cdot)$ was the lightweight decoder from SAM2 and $\mathcal{PE}(\cdot)$ was the prompt encoder.
\subsection{3D memory module: an extra path to save information from previous slices}
Processing 3D medical images in SAM2 was similar to processing video data, given the strong association between neighboring slices in 3D medical images. In our proposed network, the memory mechanism of the PMG module was extended to the 3D memory module. The 3D memory encoder encoded and stored features of previous slices and their corresponding predictions, which were stored in a 3D memory bank, denoted as
\begin{equation}
    \mathcal{M}_v = \phi(\hat{y}_v)+\mathcal{\hat{F}}_v,
    \label{eq:volumetric_memory_encoder}
\end{equation}
where $\mathcal{M}_v$ represented the encoded volumetric memory from the predictions of adjacent slices $\hat{y}_v$ and their corresponding image embeddings $\mathcal{\hat{F}}_v$. Such volumetric memory was then added to the support memory obtained from Eq~\eqref{eq:support_memory_encoder} as
\begin{equation}
    \mathcal{M}_s = [\mathcal{M}_s, \mathcal{M}_v].
    \label{eq:update_memory}
\end{equation}
The updated support memory, $\mathcal{M}_s$, was utilized in Eq~\eqref{eq:memory_attention} to generate the feature embeddings for the next slice. Through these modules, the predictions and the image embeddings of previous slices would be integrated into the support-set buffer to help compute the prediction of the current slice for 3D image segmentation. Notably, the memory bank size was set equal to the support-set size by default. During the training and inferring process, as the model propagated along the axial slices of the 3D image, the memory bank worked as a queue that accepted a new memory feature from the current axial slice and popped out an old memory feature. To improve the effectiveness of the memory-attention mechanism, we designed a memory bank that popped out the least similar memory based on the similarity score of all encoded memories and the image embeddings of the current slice.

\subsection{Pseudo-mask attention module}
The PMA module utilized pseudo-masks and image embeddings to generate the final prediction masks, with attention mechanisms playing a critical role. Transformer-based methods often exhibit slow convergence due to their reliance on global attention, which processes the entire image context indiscriminately. To address this limitation by leveraging the availability of the generated pseudo-masks, we proposed a mask-attention module inspired by Mask3D~\cite{10160590} and Mask2Former~\cite{cheng2021mask2former}. This module constrained the attention mechanism to focus on the pseudo-mask regions, enabling more effective extraction of localized features rather than the entire global image context. Mask2Former employed a two-stage decoding process where the first stage produced a coarse mask and the second stage applied this coarse mask to guide attention operation using a mask-attention module before the fine-granularity pixel decoder. Our approach extended this concept by replacing the coarse mask with the pseudo-masks generated by the PMG module. Therefore, the computational complexity was significantly reduced by focusing operations on semantically relevant regions, which was particularly important for medical images where structures of interest often occupied a small portion of the overall image. 

The PMA module consisted of three components: the SAM2 prompt encoder, the mask attention module, and the mask decoder. In the PMA module, the pseudo-masks were used for two purposes: (1) deriving bounding box prompts and (2) guiding the attention operation in the memory attention module following the Mask2Former's design. A bounding box $B_{\text{box}}$ was calculated based on the pseudo-mask  $\tilde{y_q}$, and it was used to generate prompt embeddings via the SAM2 prompt encoder. As illustrated in Fig.\ref{fig:SAM2_support}, the mask attention module fused the pseudo-mask with the image embeddings by leveraging the probabilistic map of the pseudo-mask to assign varying degrees of importance to different foreground regions. The mask attention module first resized the pseudo-mask $\tilde{y}_q$ to the same dimension as the image embeddings $\tilde{\mathcal{F}}_q$, then applied self-attention on the image embeddings $\tilde{\mathcal{F}}_q$. An element-wise product was then performed between the resized pseudo-mask and the image embeddings, effectively suppressing irrelevant regions by multiplying a near-zero probability. The module was formulated as
\begin{equation}
\begin{aligned}
    \hat{\mathcal{F}_q}={\tilde{y}_q}' \odot softmax(\frac{QK^T}{\sqrt{d}})V+\tilde{\mathcal{F}}_q,\\ 
    Q = W^Q \tilde{\mathcal{F}}_q,\quad K = W^K \tilde{\mathcal{F}}_q,\quad V = W^V \tilde{\mathcal{F}}_q
\end{aligned}
    \label{eq:mask_attention}
\end{equation}
where ${\tilde{y}_q}'$ was the resized pseudo-mask, and $W^Q$,$W^K$,$W^V$ represented the query, key, and value learnable projection matrices in the attention block, respectively.

As noted by MA-SAM~\cite{CHEN2024103310} and H-SAM~\cite{cheng2024unleashingpotentialsammedical}, the lightweight SAM decoder produced final predictions at a resolution 4 times lower than the input dimensions. The SAM2 model, employing an identical lightweight decoder structure, inherited the same limitation. To address this issue,  we adopted a hierarchical pixel decoder following the design of H-SAM. Such decoder integrated the prompt embeddings from the prompt encoder and the outputs of the mask attention module to decode the image segmentation mask, denoted as: 
\begin{equation}
    \hat{y}_q=\mathcal{D}(\hat{\mathcal{F}}_q, \mathcal{PE}(B_{\text{box}})).
    \label{eq:prediction}
\end{equation}
% denoted as
% \begin{equation}
%     B_{box} = Get\_box(\tilde{y}_q),
%     \label{eq:bbox}
% \end{equation}
% where
\label{sec:experiment}
\vspace{-0.6cm}
 \subsection{Loss function}
The training loss consisted of the loss function between the final prediction and the ground truth, and the loss function between the pseudo-mask and the final prediction. Adding the second part was to make sure the pseudo-mask generation module generated pseudo-masks close enough to the final prediction. The loss function was formulated as
\begin{equation}
  \mathcal{L} = \lambda_{dice}\mathcal{L}_{dice}(\hat{y}_q, y_q) +  \lambda_{ce}\mathcal{L}_{ce}(\hat{y}_q, y_q)+ \lambda_{KL}\mathcal{L}_{KL}(\tilde{y}_q,\hat{y}_q),
  \label{eq:loss functiton}
\end{equation}
where $\lambda_{dice}$, $\lambda_{ce}$ and $\lambda_{KL}$ were the weight hyper-parameters for each loss function, $\mathcal{L}_{dice}$ denoted the dice loss function, $\mathcal{L}_{ce}$ denoted the binary cross entropy loss function, $\mathcal{L}_{KL}$ denoted the KL divergence loss function, $\tilde{y}_q$ denoted the pseudo-mask from the PMG module, $\hat{y}_q$ denoted the final prediction, and $y_q$ denoted the ground truth.

\begin{figure*}
  \centering
\includegraphics[width=0.93\linewidth]{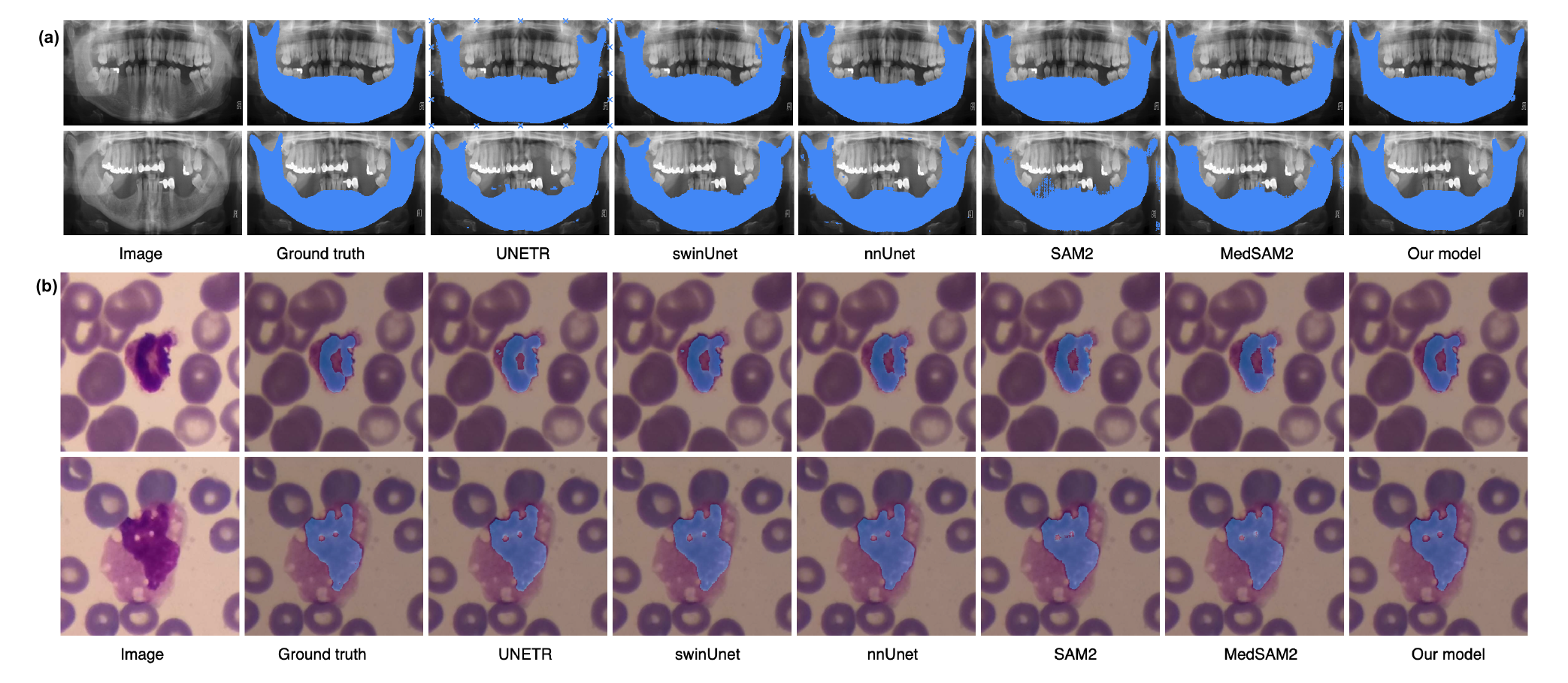}
      \caption{\small{Examples of segmentation results from the Pandental and WBC datasets. For each sub-figure, the first column shows the image, and subsequent columns present results from ground truth and 6 comparison methods. The two rows represent two different cases.  (a) Segmentation results of bones (labelled in blue) from the Pandental dataset. (b) Segmentation results of the nucleus (labelled in blue) from the WBC dataset.}}
\label{fig:2Dvisualization_2}
\end{figure*}

\begin{figure*}
  \centering
  \includegraphics[width=0.93\linewidth]{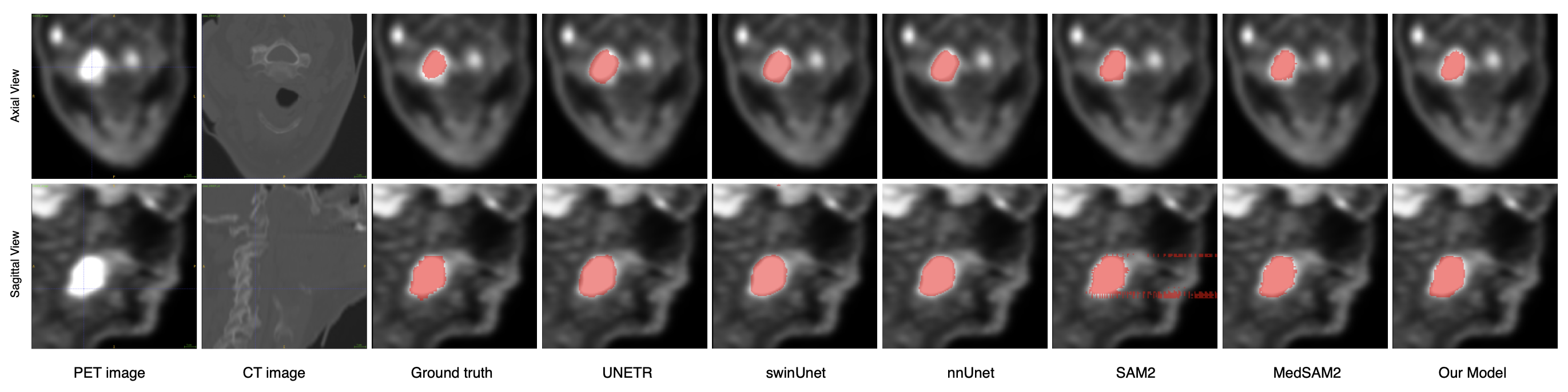}
  \caption{\small{Examples of segmentation results from the PET/CT 3D dataset. The two rows represent the sagittal and axial views. The first column shows the PET image, the second column shows the CT image, and subsequent columns present segmentation results (labelled in pink) from ground truth and 6 comparison methods.}}
  \label{fig:petct visualization}
\end{figure*}

\begin{figure*}
  \centering
  \includegraphics[width=0.93\linewidth]{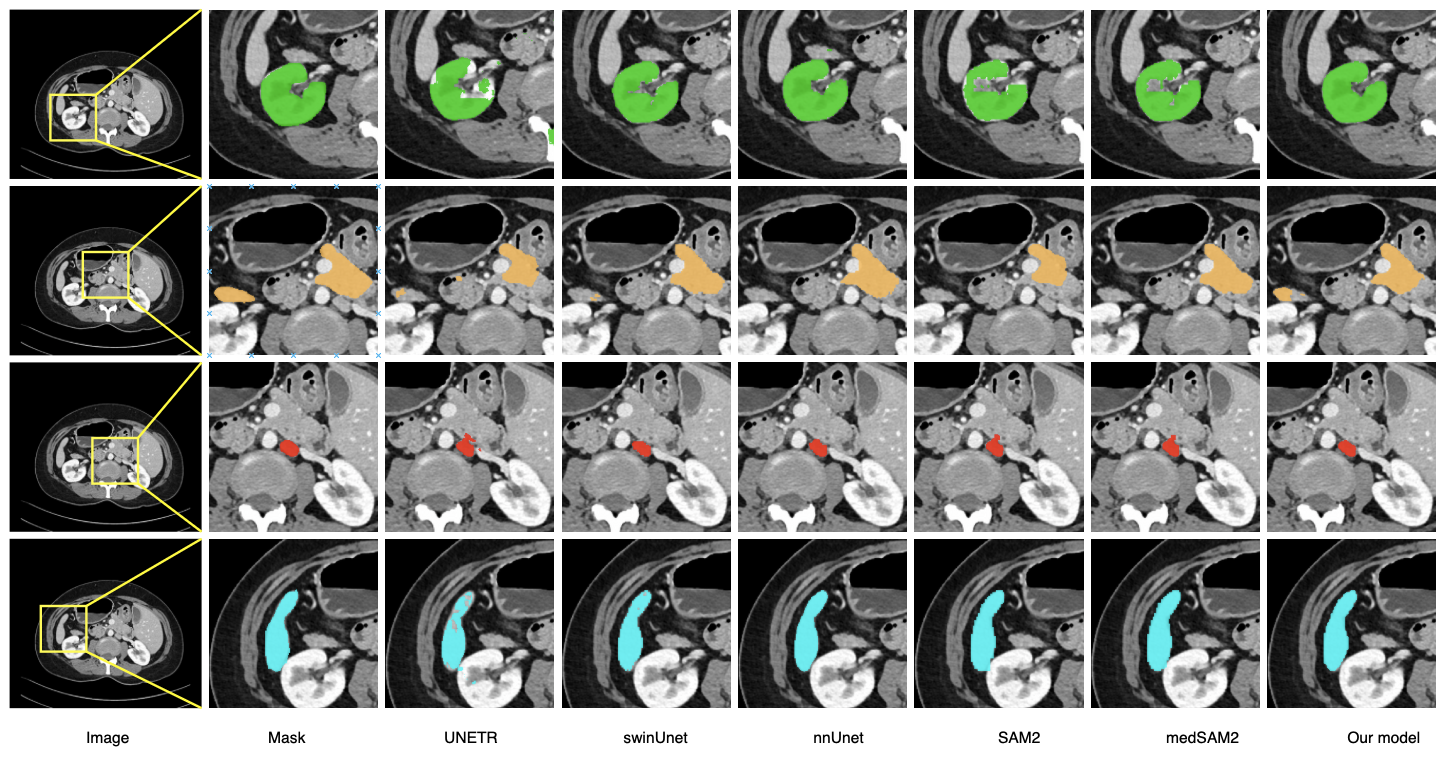}
  \caption{\small{Examples of segmentation results from the AMOS22 3D dataset. Different organs were labeled in different colors, including left kidney (in green), spleen (in cyan), postcava (in orange), and pancreas (in red). The first column shows the image, and subsequent columns present results from ground truth and 6 comparison methods. The different rows came from the same CT slice. }}
  \label{fig:Amos visualization}
\end{figure*}
\begin{figure*}
  \centering
  \includegraphics[width=0.93\linewidth]{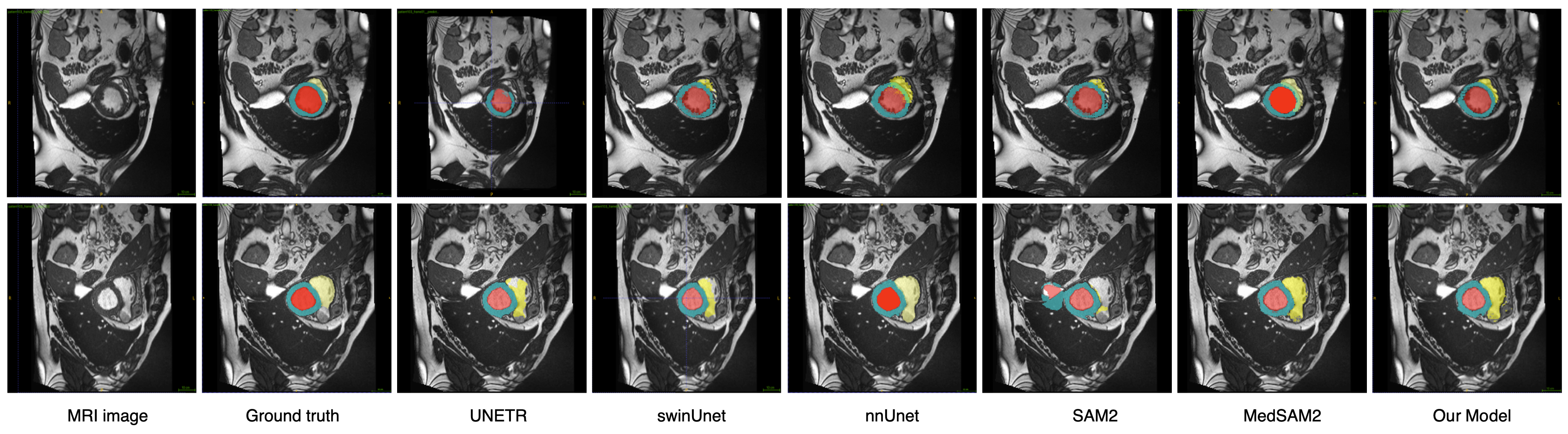}
  \caption{\small{Examples of segmentation results from the ACDC 3D dataset. Left ventricle, right ventricle, and myocardium were labelled in red, yellow, and cyan, respectively. The first column shows the image, and subsequent columns present results from ground truth and 6 comparison methods. The two rows represent two different cases. }}
  \label{fig:ACDC visualization}
\end{figure*}

\begin{table}
  \centering
  \footnotesize
  \begin{tabular}{c c c c c c}
    \toprule
     Model & REFUGE & PanDental & WBC & CAMUS & BUSI\\
    \midrule
    UNETR   & 0.752  & 0.906 & 0.939 & 0.803 & 0.554\\
    SwinUnet   & 0.864  & 0.904 & 0.906 & 0.920 & 0.733\\
    nnUnet     & 0.849  & 0.929 & 0.951 & 0.925 & 0.751\\
    SAM2       & 0.753  & 0.854 & 0.627 & 0.830 & 0.577\\
    medSAM2    & 0.805  & 0.941 & 0.951 & 0.919 & 0.627\\
    Our model  & \textbf{0.865}  & \textbf{0.969} & \textbf{0.976} & \textbf{0.932} & \textbf{0.767}\\
    \bottomrule
  \end{tabular}
  \caption{Quantitative results: Dice scores on 2D datasets}
  \label{tab:2dresults}
\end{table}

\section{Experiment}
\subsection{Datasets}
Multiple public datasets were used to evaluate performance of the proposed model on both 2D and 3D  tasks. The selected datasets covered a diverse range of medical imaging modalities and anatomical structures, with details  below.
\subsubsection{2D datasets}
5 different 2D datasets were used, including:  (1) White blood cell (WBC) Dataset~\cite{kouzehkanan2022large} was a collection of microscopic images of white blood cells, used for segmentation and classification tasks. It contained 12,500 images with a resolution of 320×320 pixels, capturing various types of WBCs.  (2) Panoramic dental dataset contained high-resolution panoramic dental radiographs annotated for tooth segmentation and bone segmentation~\cite{abdi2015automatic}.  It consisted of over 1,000 images with a resolution of approximately 2,000×1,000 pixels. (3) CAMUS dataset~\cite{Leclerc2019} was a cardiac ultrasound dataset with manual segmentation of the left ventricle, myocardium, and atrium. It contained 1,800 echocardiographic images with a resolution of 512×512 pixels. (4) BUSI dataset~\cite{ALDHABYANI2020104863} was a breast ultrasound dataset containing images with tumor annotations for segmentation. It included 780 images with a resolution of 500×500 pixels. (5) REFUGE dataset~\cite{ORLANDO2020101570} contained retinal fundus images designed for the segmentation of the optic disc and cup, which helped to detect glaucoma. The images had a resolution of 2,120×2,120 pixels.
\subsubsection{3D datasets} 3 different 3D datasets were utilized. (1) Head-and-neck PET/CT dataset~\cite{https://doi.org/10.1002/mp.16703} contained 100 3D PET images with a resolution of 124×124 pixels per slice, supporting the segmentation of tumors and organs at risk for radiation therapy planning. (2) The Automated Cardiac Diagnosis Challenge (ACDC) dataset ~\cite{8360453} comprised cardiac MR images with segmentation annotations for the left and right ventricles and myocardium. It contained 150 cases with a resolution of 256×256 pixels for each slice. (3) Abdominal Multi-Organ Segmentation 2022 (AMOS22) dataset ~\cite{ji2022amos} was a multi-organ segmentation dataset based on abdominal CT scans, providing annotations for various organs. It included 500 cases with image resolutions ranging from 512×512 to 512×1024, covering 15 abdominal organs. \\
\begin{table}[t]
  \centering
  \footnotesize
  \begin{tabular}{c c c c}
    \toprule
     Model & AMOS\_CT  & AMOS\_MRI & PET/CT\\
    \midrule
    UNETR       & 0.818 & 0.579 & 0.694 \\
    SwinUnet    & 0.888 & 0.575 & 0.732 \\
    nnUnet      & 0.863 & 0.676 & 0.724 \\
    SAM2        & 0.699 & 0.560 & 0.682\\
    medSAM2     & 0.859 & 0.654 & 0.709\\
    Our model   & \textbf{0.903} & \textbf{0.715} & \textbf{0.745}\\
    \bottomrule
  \end{tabular}
  \caption{Quantitative results: Dice scores on the AMOS22 and PET/CT 3D datasets.}
  \label{tab:AMOSresults}
\end{table}

\begin{table*}[htp]
  \centering
  \footnotesize
  \begin{tabular}{c c c c c c c c c c c}
    \toprule
     Model & Spleen & Right Kidney & Left Kidney & Gallbladder & Esophagus & Liver & Stomach & Aorta & Post Cava & Pancreas\\
    \midrule
    UNETR       & 0.927 & 0.885 & 0.901 & 0.665 & 0.733 & 0.941 & 0.787 & 0.914 & 0.682 & 0.745\\
    SwinUnet    & 0.955 & 0.938 & 0.945 & 0.773 & 0.831 & 0.960 & 0.889 & 0.947 & 0.749 & 0.849\\
    nnUnet      & 0.963 & \textbf{0.953} & 0.963 & 0.815 & \textbf{0.857} & 0.971 & \textbf{0.908} & 0.954 & \textbf{0.796} & \textbf{0.874}\\
    SAM2        & 0.857 & 0.855 & 0.857 & 0.800 & 0.643 & 0.811 & 0.759 & 0.842 & 0.637 & 0.538\\
    medSAM2     & 0.956 & 0.940 & 0.941 & 0.792 & 0.725 & 0.947 & 0.758 & 0.831 & 0.772 & 0.798\\
    Our model   & \textbf{0.966} & 0.952 & \textbf{0.969} & \textbf{0.882} & 0.852 & \textbf{0.974} & 0.888 & \textbf{0.962} & 0.793 & 0.851\\
    \bottomrule
  \end{tabular}
  \caption{Quantitative results: Dice scores for 10 organs on the AMOS22 3D dataset.}
  \label{tab:AMOSorganwise}
\end{table*}
\begin{table}[t]
  \centering
  \footnotesize
  \begin{tabular}{c c c c c}
    \toprule
     Model & Average & left heart & right heart & myocardium\\
    \midrule
    UNETR       & 0.866 & 0.940 & 0.853 & 0.865\\
    SwinUnet    & 0.900 & \textbf{0.958} & 0.885 & 0.856\\
    nnUnet      & 0.916 & 0.954 & 0.902 & \textbf{0.892}\\
    SAM2        & 0.539 & 0.695 & 0.444 & 0.479\\
    MedSAM2     & 0.810 & 0.763 & 0.791 & 0.876\\
    Our model   & \textbf{0.917} & 0.954 & \textbf{0.911} & 0.885\\
    \bottomrule
  \end{tabular}
  \caption{Quantitative results: Dice scores on the ACDC 3D dataset.}
  \label{tab:ACDCresults}
\end{table}

\vspace{-0.3cm}
\subsection{Evaluation and implementation details}
Segmentation performance was evaluated using the Dice similarity coefficient (Dice). All models were trained using the Adam optimizer and NVIDIA A100 GPUs, with a single A100 GPU used for 2D datasets and four A100 GPUs for 3D datasets. Hyperparameters, including learning rates and batch sizes, were tuned for optimal performance across different datasets. For 3D reference methods such as swinUnet and UNETR, which only accepted 3D images with a minimal patch size requirement of 32x32x32, each sample of the 2D dataset was repeated 32 times and stacked as a 3D sample as input. 

\section{Results}
\vspace{0.2cm}
\subsection{2D segmentation results}
Table~\ref{tab:2dresults} presents a quantitative comparison of the proposed model with several baseline methods, including UNETR~\cite{hatamizadeh2022unetr}, SwinUnet~\cite{hatamizadeh2022swinunetrswintransformers}, nnUNet~\cite{Isensee2021}, SAM2~\cite{ravi2024sam2segmentimages}, and medSAM2~\cite{zhu2024medicalsam2segment}, on REFUGE, Panoramic Dental, WBC, CAMUS, and BUSI datasets. The proposed model consistently achieved superior performance across all evaluated 2D tasks. Note that for SAM-based methods, a random click was used as the prompt. Qualitative examples of the segmentation outcomes are illustrated in Fig.\ref{fig:2Dvisualization} and Fig.\ref{fig:2Dvisualization_2}, further demonstrating the robustness and generalization capabilities of the proposed model in diverse 2D image segmentation scenarios.

\subsection{3D segmentation results}
Table~\ref{tab:AMOSresults} summarizes the segmentation performance of various models on the PET/CT dataset~\cite{yu2024pet}, which shows that the proposed model achieved the best performance compared to other reference methods.  Additionally, Table~\ref{tab:AMOSresults} compares Dice scores for CT and MRI images from the AMOS22 dataset, while Table~\ref{tab:AMOSorganwise} provides detailed organ-wise segmentation results specifically for CT images.  The proposed model achieved the highest overall Dice scores on both CT and MRI images from the AMOS22 dataset, excelling in 5 out of 10 organs in the organ-wise segmentation evaluation, thus highlighting its effectiveness for multi-organ segmentation tasks. Table~\ref{tab:ACDCresults} presents Dice scores for the ACDC dataset, including both average and target-specific  scores. Although the proposed model exhibited slightly lower performance in segmenting the left ventricle and myocardium, it achieved the highest overall Dice score, underscoring its competitive performance on 3D cardiac MRI segmentation. Qualitative segmentation examples are provided in Fig.~\ref{fig:petct visualization} for the PET/CT dataset, Fig.~\ref{fig:Amos visualization} for the AMOS22 dataset, and Fig.~\ref{fig:ACDC visualization} for the ACDC dataset. These qualitative examples align with the quantitative findings, further illustrating the model's capability in generating accurate and consistent segmentation masks.
\begin{table}
  \centering
  \footnotesize
  \begin{tabular}{c c c c c c}
    \toprule
     Experiment & 3DM & PMG  & PMA & left kidney & right kidney\\
    \midrule
    (a)    &  &  & \checkmark & 0.936 & 0.928\\
    (b)    &  & \checkmark & \checkmark & 0.947 & 0.942\\
    (c)    & \checkmark & \checkmark & \checkmark & \textbf{0.966} & \textbf{0.952}\\
    \bottomrule
  \end{tabular}
  \caption{Ablation study on 3DM, PMG, PMA modules on the AMOS22 3D dataset. Performance was reported in Dice scores for left and right kidneys.}
  \label{tab:abl_components}
\end{table}
\begin{table*}
  \centering
  \footnotesize
  \begin{tabular}{c cc cc cc cc}
    \toprule
     Support Size & \multicolumn{2}{c}{Pandental} & \multicolumn{2}{c}{CAMUS} & \multicolumn{2}{c}{BUSI} & \multicolumn{2}{c}{WBC} \\
    \midrule
    & Dice & IoU & Dice & IoU & Dice & IoU & Dice & IoU \\
    \midrule
    1  & 0.9660 & 0.9370 & 0.9302 & 0.8710 & 0.7470 & 0.6706 & 0.9743 & 0.9546 \\
    2  & 0.9694 & 0.9406 & 0.9324 & 0.8751 & 0.7699 & 0.7010 & 0.9745 & 0.9552 \\
    4  & 0.9695 & 0.9409 & 0.9323 & 0.8749 & 0.7673 & 0.6925 & 0.9760 & 0.9579 \\
    8  & 0.9685 & 0.9391 & 0.9327 & 0.8755 & 0.7102 & 0.6340 & 0.9812 & 0.9634 \\
    16 & \textbf{0.9721} & \textbf{0.9457} & \textbf{0.9342} & \textbf{0.8777} & \textbf{0.7730} & \textbf{0.6924} & \textbf{0.9825} & \textbf{0.9659} \\
    \bottomrule
  \end{tabular}
  \caption{Ablation study on different support-set size. Performance was reported in Dice scores and IoU across different  datasets.}
  \label{tab:ablation_support}
\end{table*}
\subsection{Ablation studies}
\label{sec:ablation}
We first performed ablation studies to evaluate the effectiveness of each component. Furthermore, additional ablation analyses were performed to investigate the impact of key fine-tuning settings, specifically: (1) support-set size,  and (2) SAM2's pretrained weights.
\subsubsection{Effectiveness of each component}
To evaluate the contribution of individual modules to the overall segmentation performance, an ablation study was conducted on the AMOS22 3D dataset, selectively disabling each component in turn. Table~\ref{tab:abl_components} presents the experimental results specifically for the left and right kidneys. Initially, both the 3DM and PMG modules, which were responsible for integrating contextual information from the support set and previous slices, were disabled, effectively reducing the model to a baseline SAM2 fine-tuned with a customized segmentation head. Experiment (a) revealed superior performance even at this baseline compared to advanced segmentation methods, highlighting the robustness of the SAM2 backbone. In experiment (b), only the 3DM module was disabled, and comparison to experiment (a) indicated notable improvements when the PMG module was enabled, validating the benefit of pseudo-mask guidance from the support set. Further comparison between experiments (b) and (c), where the 3DM module was subsequently enabled, demonstrated additional performance gains, emphasizing the importance of temporal memory from adjacent slices in the pseudo-mask generation process.
\subsubsection{Support-set size}
 To examine how the support-set size affected performance, ablation experiments were designed by keeping the training set fixed and varying the support-set size. Specifically, a subset of training samples was duplicated to serve as the support set, ensuring that the training distribution remained unchanged. As shown in Table~\ref{tab:ablation_support}, enlarging the support set improved average Dice and Intersection over Union (IoU) scores, due to enhanced information provided by the larger support sets. This richer information resulted in more precise pseudo-mask generation through the support memory attention module. Nevertheless, increasing the support-set size also introduced additional computational overhead. Consequently, a support-set size of 4 was chosen as the default to achieve an optimal balance between performance and computational efficiency.
 \subsubsection{SAM2 pretrained weights}
Ablation experiments were also conducted to evaluate the impact of initializing the model with the released weights of SAM2 versus training from scratch. Although SAM2 was originally trained on natural images, Table~\ref{tab:ablation_weight} demonstrates that the pretrained weights provided substantial benefits when transferred to medical domain, including better convergence and stronger zero-shot performance. These findings highlighted the strong generalizability and robustness of SAM2's learned representations, even when applied to domain-shifted tasks such as medical image segmentation.

\begin{table}
  \centering
  \footnotesize
  \begin{tabular}{c cc cc}
    \toprule
    Dataset & with SAM2 weights & without SAM2 weights\\
    \midrule
    WBC        & 0.961  & 0.695 \\
    Pandental  & 0.951 & 0.801 \\
    REFUGE     & 0.865  & 0.434 \\
    BUSI       & 0.909 & 0.158 \\
    CAMUS      & 0.930 & 0.427 \\
    \bottomrule
  \end{tabular}
  \caption{Ablation study on the effect of adopting SAM2's pretrained weight.  Performance was reported in Dice scores across different datasets.}
  \label{tab:ablation_weight}
\end{table}

\section{Discussion}
\label{sec:conclusion}
In this work, we introduced a novel adaptation of SAM2 for medical image segmentation, leveraging the memory mechanism of SAM2 and in-context learning to address two major challenges of applying SAM2 model to medical image segmentation tasks: the reliance on high-quality human-guided prompts and the domain shift problem caused by SAM2’s training on natural images. By integrating a cross-attention module to generate pseudo-masks based on support sets, the proposed model automated the prompt generation and enhanced segmentation performance across various modalities, including fundus photography, X-ray, CT, MRI, PET, and ultrasound. The proposed method demonstrated significant improvements over other reference methods. Overall, the proposed approach represented a step toward automating and improving medical image segmentation using foundation models such as SAM2.

While the proposed method demonstrated promising performance, it also had certain limitations. First, the quality of the pseudo-masks was dependent on the quality of the support set used, which might affect the model's robustness in scenarios with limited labeled data. Second, the strategy of applying the proposed model on the 3D dataset was to treat 3D images as videos, which could be considered as a temporal sequence of slices along one direction. However, this unidirectional propagation, from the first to the last slice, leveraged only half of the adjacent contextual information when predicting each slice, potentially limiting segmentation accuracy. Exploring strategies such as bidirectional propagation to process 3D datasets remains an area for future investigation. Additionally, further refining the pseudo-mask generation process and extending the model to efficiently process multi-modal imaging, e.g., utilizing both PET and CT images as inputs for the PET/CT 3D dataset, deserves further investigations.

% In summary, our research directions included and enhancing the efficiency of 3D segmentation by using different propagation strategies. With continued advancements, the integration of foundation models in medical imaging has the potential to revolutionize automated diagnosis and treatment planning.

\section{Conclusion}
This study proposed a novel foundation model for medical image segmentation built upon the SAM2 architecture. By incorporating in-context learning and pseudo-mask driven prompting, the proposed model effectively addressed the challenges of adapting SAM2 to the medical domain and eliminated the need for human guidance. Evaluations across a diverse set of 2D and 3D datasets demonstrated consistent improvements of the proposed framework over both traditional and SAM-based segmentation methods.


% % use section* for acknowledgment
% \ifCLASSOPTIONcompsoc
%   % The Computer Society usually uses the plural form
%   \section*{Acknowledgments}
% \else
%   % regular IEEE prefers the singular form
% \fi




%\bibliographystyle{IEEEtran}

\bibliographystyle{IEEEtran}
\bibliography{IEEEabrv, bib/ref}

\end{document}